\documentclass[letterpaper, 10 pt, conference]{ieeeconf}  % Comment this line out if you need a4paper

\usepackage{cite}
\usepackage{amsmath,amssymb,amsfonts}
\usepackage{algorithmic}
\usepackage{graphicx}
\usepackage{textcomp}
\usepackage{xcolor}
\usepackage[labelfont=bf,font=small]{caption}
\usepackage{hyperref}

\usepackage{booktabs}

\IEEEoverridecommandlockouts                              % This command is only needed if 
\title{\LARGE \bf
 A Brief Survey on Leveraging Large Scale Vision Models \\for Enhanced Robot Grasping}

\author{Abhi Kamboj and Katie Driggs-Campbell}

\begin{document}

\maketitle
\thispagestyle{empty}
\pagestyle{empty}

%%%%%%%%%%%%%%%%%%%%%%%%%%%%%%%%%%%%%%%%%%%%%%%%%%%%%%%%%%%%%%%%%%%%%%%%%%%%%%%%
\begin{abstract}
Robotic grasping presents a difficult motor task in real-world scenarios, constituting a major hurdle to the deployment of capable robots across various industries. 
Notably, the scarcity of data makes grasping particularly challenging for learned models. 
Recent advancements in computer vision have witnessed a growth of successful unsupervised training mechanisms predicated on massive amounts of data sourced from the internet and now nearly all prominent models leverage pretrained backbone networks. 
Against this backdrop, we begin to investigate the potential benefits of large-scale visual pretraining in enhancing robot grasping performance. 
This preliminary literature review sheds light on critical challenges and delineates prospective directions for future research in visual pretraining for robotic manipulation.

\end{abstract}

%%%%%%%%%%%%%%%%%%%%%%%%%%%%%%%%%%%%%%%%%%%%%%%%%%%%%%%%%%%%%%%%%%%%%%%%%%%%%%%%
\section{Introduction}
Robot grasping has been applied in various industries, including agriculture, manufacturing, and automation. 
For example, Kim et al.~\cite{Kim2019} proposed the use of robots equipped with sophisticated grasping algorithms for crop harvesting, increasing efficiency and reducing labor costs. 
In the manufacturing industry, Lee et al.~\cite{Lee2018} presented a study on the use of robotic grasping in the assembly of parts, improving production speed and reducing errors. 
Jain et al.~\cite{Jain2016} investigated the application of robotic grasping in material handling and packaging, demonstrating improved efficiency and reduced operational costs.  
Furthermore, grasping robots can be used in hazardous environments, such as those with toxic materials, to minimize the risk to human workers.
These applications demonstrate the versatility and practicality of robot grasping and its potential to revolutionize many industries.

Robot grasp planning can broadly be divided into 3 main subtasks: object localization, object pose estimation, and grasp estimation \cite{du2021vision}.
Robotic grasping approaches first require knowledge of the object's location in a scene, which can involve object detection, segmentation, or tracking.
Then for the robot to manipulate the object it performs 6D pose estimation (3D translation and 3D rotation) transforming the object from the object coordinate to the camera coordinate.
Finally, grasp estimation estimates the 6D gripper pose in the camera coordinate, often having to use prior knowledge of the object as well as its location and pose.
These distinct tasks are only one view of robot grasp planning and many works explore performing two steps jointly or performing all three at once.

\section{Challenges}
The deployment of robot grasping is a challenging task that has not yet reached full capability in most instances. 
We discuss two related challenges that visual pretraining may be able to help overcome. 
The first is insufficient information about the object, and the second is insufficient training data.

Insufficient information about the object presents a major challenge as it is difficult to predict and execute the optimal grasp for an object without knowing its full geometry or dynamics. \cite{du2021vision}.
Even with multi-view or depth camera information, objects are likely to be partially occluded, making it difficult to predict the optimal grasp \cite{berenson2008grasp, kasaei2023mvgrasp}. 
Accounting for an object's shape is particularly difficult when adapting to novel or transparent objects in which case the system may have to generalize prior knowledge to reason about an effective grasp \cite{saxena2008robotic, sajjan2020clear}.

In addition, robotics struggles to obtain sufficient training data compared to generic computer vision and detection models. 
Unlike generic computer vision and detection models, robotics often relies on trial and error from its own experiences. 
Thus the robot needs to spend a lot of time training and exploring different scenarios to be able to perform well and generalize to unseen scenarios.
Although simulation can be used to generate additional data, there is often a 'reality gap' \cite{tremblay2018deep} or 'sim2real' \cite{kim2022ipc} transfer issue that needs to be addressed.

The lack of labeled training data implies that the robot is not able to sufficiently learn object representations. The lack of data precludes robotics models from being able to advance rapidly and scale as other state-of-the-art computer vision models have done with large quantities of data. 
For example, contrastive language-image pretraining (CLIP) is a self-supervised method of image pretraining a model that revolutionized computer vision.
The original paper \cite{radford2021learning}, pulls hundreds of millions of images from online with caption labels and uses contrastive learning to learn image representations and their correlations with language. 
Robotics does not have such a resource where it can get a massive number of episodes to learn a control or decision policy. 
\begin{figure*}[t]
\centerline{\includegraphics[width=\textwidth]{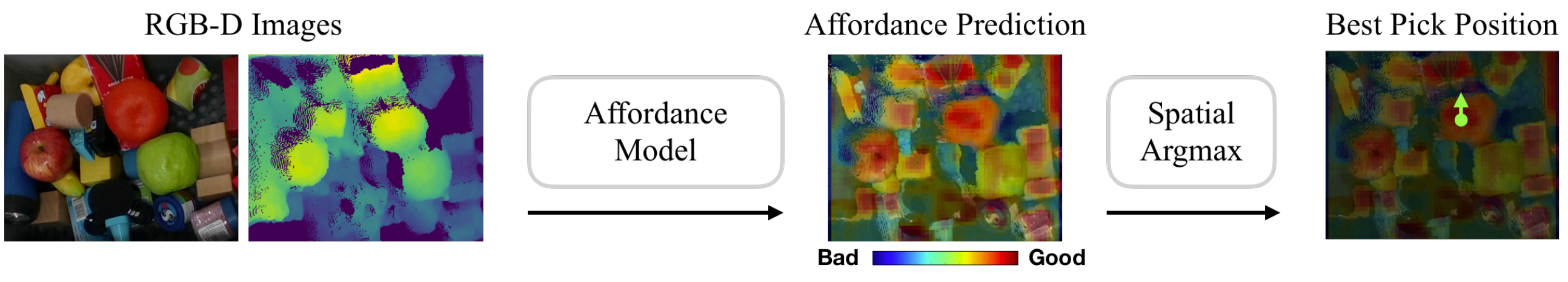}}
\caption{A visualization of the affordance prediction network in \cite{yen2020learning}. An RGBD image is used as input, and the output is a heatmap indicating which locations are "Good" or "Bad" for the gripper or suction cup to pick up.}
\label{affordance_fig}
\vspace{-10pt}
\end{figure*}

\section{Visual Pretraining in Robotics}
Many recent works have attempted to leverage visual pretraining for robotic manipulation, to address the challenges mentioned in the previous section \cite{xiao2022masked,yen2020learning,radosavovic2022real,nair2022r3m}. 

Yen-Chen et al. \cite{yen2020learning} notice that the concept of affordance maps in visual representations may correlate with the output of standard vision models.
An affordance map in robot manipulation usually refers to a heat map predicting locations in the image where the robot has a higher probability of executing a successful grasp (See Fig \ref{affordance_fig}).
Yen-Chen et al. train vision models for various passive vision tasks (e.g. normal estimation, object detection, segmentation, and edge detection) and use that model as an initialization for an affordance map estimation network in 2D.
They then test the affordance maps with predefined suction and parallel-jaw grasping primitives.
The results show that first learning visual predictions and then learning affordance models from the same weights gives a strong initial policy, significantly reduces the number of interactions required for learning a policy (improves sample efficiency), and improves speed and performance for learning manipulation in a new environment and unseen objects.

Radosavovic et al. \cite{radosavovic2022real} pretrain visual representations on millions of egocentric hand object interaction videos using a masked autoencoder (MAE) and then freeze the encoder, append a controller network and train a robotic manipulation policy using behavior cloning on human demonstrations. 
Their intuition was that by learning image representations through masking parts of the image, the model will learn useful real-world properties of objects that can aid it in learning to manipulate those objects.
Their results indicate that MAE pretraining improves sample efficiency, transferability to multiple tasks (e.g push, pull, pick), and performance accuracy in complex tasks implying that MAE-trained visual representations capture spatial information helpful for robotics tasks.

Nair et al. \cite{nair2022r3m} add language into the mix and use time contrastive learning with video language alignment on large diverse human video datasets (e.g. Ego4D \cite{Ego4D2022CVPR}).
They show that pretraining a visual representation improves robot manipulation task performance significantly, i.e. they learn to pick objects in just 20 demonstrations.

\section{Future Directions}
Affordance estimation elegantly performs object localization, pose estimation, and grasp prediction in one step through visual representations. 
Given the vast quantity of data that can pretrain an affordance estimation model through various related passive visual tasks, promising future research directions may involve more robust visual affordance representations.
For example, \cite{mandikal2021learning} learns difficult dexterous grasping using 3D data, however, is limited to a labeled 3D dataset. 
Extending the affordance estimation through pretraining work of \cite{yen2020learning} to 3D may allow for more complicated grasps.
Mapping learned 2D visual representations to 3D may involve simulation, multi-view cameras, or keypoint lifting to infer 3D geometry from 2D data. 

Another interesting direction would be to apply pretraining to some aspect of reinforcement learning (RL).
For example, since data efficiency is often a bottleneck in RL, a pretrained vision model could be used for guided exploration similar to the vision-based exploration model for grasping used in \cite{kalashnikov2018qt,wang2021roll}.
Alternatively, a pretrained affordance estimation model can provide a potential reward function or state value estimate to create training samples that were not collected in reality. 
Using a hand object interaction dataset (e.g. \cite{cao2021reconstructing,hampali2020honnotate}), the hand can be tracked as an agent, and affordance estimation can be used as a reward function, allowing the videos to be used as episodes to train an RL policy. 

Finally, there is room to explore large-scale visual pretraining in end-to-end robot grasping \cite{jang2017end, wu2020grasp, ainetter2021end,zeng2018learning}.
As opposed to initializing an affordance prediction model, training it, and then training a separate manipulation policy, attempting to train one network to predict and perform grasp estimation through a pretrained model may show better results. 
Intuitively, leveraging the strong pretrained visual representations closer to the robotic manipulation task at hand may perform better since the information provided by the representations could have been lost or diminished in a longer multi-step process.

\section{Conclusion}
In this work, we present a targeted overview of robot grasping methods related to visual pretraining. 
Visual pretraining could potentially aid in overcoming two main obstacles that prevent robots from achieving effective grasping. 
One is an insufficient visual understanding of objects, and the second is a lack of training data.
Visual pretraining has been crucial in advancing computer vision, and utilizing this technique is a highly promising approach for enhancing robotic grasping to the extent that it can be adopted in more practical and diverse applications.


%%%%%%%%%%%%%%%%%%%%%%%%%%%%%%%%%%%%%%%%%%%%%%%%%%%%%%%%%%%%%%%%%%%%%%%%%%%%%%%%

%%%%%%%%%%%%%%%%%%%%%%%%%%%%%%%%%%%%%%%%%%%%%%%%%%%%%%%%%%%%%%%%%%%%%%%%%%%%%%%%

%%%%%%%%%%%%%%%%%%%%%%%%%%%%%%%%%%%%%%%%%%%%%%%%%%%%%%%%%%%%%%%%%%%%%%%%%%%%%%%%

%%%%%%%%%%%%%%%%%%%%%%%%%%%%%%%%%%%%%%%%%%%%%%%%%%%%%%%%%%%%%%%%%%%%%%%%%%%%%%%%

\bibliographystyle{IEEEtran}
\bibliography{IEEEabrv,BibFile}

\end{document}